\pdfoutput=1

\documentclass[11pt]{article}

\usepackage{ACL2023}

\newcommand{\DylanRemark}[1]{\textbf{\textcolor{red}{DylanRemark: #1}}}
\usepackage{times}
\usepackage{latexsym}
\usepackage{graphicx}
\usepackage{subcaption}
\usepackage{booktabs}
\usepackage[T1]{fontenc}

\usepackage[utf8]{inputenc}
\usepackage[subtle]{savetrees}
\usepackage{microtype}
\usepackage{amsmath}

\usepackage{inconsolata}
\usepackage{enumitem}

\title{Instruction Diversity Drives Generalization To Unseen Tasks}

\author{Dylan Zhang \\
    University of Illinois \\Urbana-Champaign \\
  \texttt{shizhuo2@illinois.edu} \\\And
  Justin Wang \\
        University of Illinois \\Urbana-Champaign \\
  \texttt{jw93@illinois.edu} \\
  \\\And
  Francois Charton \\
        Meta AI Research \\
  \texttt{fcharton@meta.com} \\
  }
\usepackage{xcolor} 

\begin{document}
\maketitle
\begin{abstract}
Instruction tuning -- fine-tuning a large language model (LLM) on pairs of instructions and desired outcomes -- is an approach that enables pre-trained language models to perform real-world tasks and follow human instructions. Its practical success depends on the model learning a broader set of instructions than those it was trained on. Yet the factors that determine model generalization to such \emph{unseen tasks} are not well understood.
In this paper, we experiment with string rewrites, a symbolic task that serves as a building block for Turing complete Markov algorithms while allowing experimental control of ``inputs'' and ``instructions''. We investigate the trade-off between the number of instructions the model is trained on and the number of training samples provided for each instruction and observe that the diversity of the instruction set determines generalization. Generalization emerges once a diverse enough set of tasks is provided, even though very few examples are provided for each task. Instruction diversity also ensures robustness with respect to non-uniform distributions of instructions in the training set.



\end{abstract}

\section{Introduction}

The rapid advance of large language models (LLM) is one of the most exciting recent developments in artificial intelligence. LLM, pre-trained on large text corpora, can be fine-tuned to achieve high performance over a broad set of tasks, ranging from natural language understanding to logical and mathematical reasoning and programming. 


\emph{Instruction tuning} -- training models on pairs of instructions and desired outcomes -- emerged as an approach to adapt language models pre-trained over text corpus with a next-token-prediction objective to solving certain problems with its knowledge and reasoning capabilities. Through instruction-tuning, LLMs are expected to learn to perform a broad set of tasks to solve a large number of real-world problems and seamlessly interact with humans. The success of instruction tuning is therefore conditioned by the model's ability to learn to generalize to \emph{unseen tasks} -- instructions that were not seen during training.

Whereas large instruction-tuning datasets of increasing quality have been proposed in recent years, there are comparatively few systematic studies of the key factors that enable instruction-tuned models to generalize to unseen tasks. 
Intuitively, several potential factors can improve finetuning: larger training samples, a greater diversity of instructions, and better annotation quality.
However,  the real-world instruction-following datasets lack control over each factor, failing to answer the question in a systematic and principled way.

In this paper, we study task generalization in instruction-following on a simple symbolic task: string rewrites. Symbolic tasks allow us to gain finer control over the data. Also, this string-rewriting setup allows us to separate between the ``inputs'' and ``instructions'' and vary each independently. 

Our string-rewriting tasks are inspired by Markov algorithm~\cite{markov54}, which is a Turing-complete computation model. A model that can generalize string rewrites can therefore be turned into a general model of computation. Thus, the symbolic task we adopt has considerable generality over the tasks LLM performs.
We conclude that 1) {\bf instruction diversity is the enabling factor for generalization}. Models generalize once they are trained on enough different instructions, even if the number of examples per instruction is small 2) {\bf Semantic diversity of rules is also important in addition to the number of instructions} and
3) {\bf non-uniformity in distribution can affect generalization, but a diverse enough set of instructions provides robustness}.

\section{Related Work}

Both the size and quality of fine-tuning are important~\cite{chung2022scaling,iyer2022opt,wang2023far}.
High-quality datasets for instruction tuning can be collated using humans annotators~\cite{khashabi2020unifiedqa,ye2021crossfit,sanh2022multitask,wang2022supernaturalinstructions,longpre2023flan,DatabricksBlog2023DollyV2,köpf2023openassistant}, but their size is constrained by the cost of annotation. Alternative methods, using examples distilled from larger, more powerful language models have been  proposed~\cite{wang2023selfinstruct,honovich2022unnatural,alpaca,peng2023instruction,vicuna2023,xu2023wizardlm,köksal2023longform,kim2023cot}. They allow for the larger training set, in exchange for potentially lower quality. 


On generalization to unseen instruction, previous research has shown that data quality matters more than quantity~\cite{zhou2023lima}, and other works pointed out the importance of consistency in format~\cite{liang2024formatconsistency} and mixing tasks from different categories~\cite{longpre2023flan,iyer2022opt,bukharin2024data}. 

\section{Markov Algorithms and rewrite rules}

Markov algorithms~\citep{markov54} process sequences of letters on a fixed alphabet $\Sigma = \{\sigma_1, ...,\sigma_K \} $. An algorithm is an ordered sequence of rewrite rules $I = \{(x_i \rightarrow y_i)\:i=1,2...,|I|\}$, with $s_i$ and $y_i$ words over an extension of $\Sigma$: $\Sigma' = \Sigma + \{\alpha_1, ...,\alpha_n \} + \{\cdot\}$. To apply the algorithm to a given sequence $z$, rules are considered in order, and the first applicable rule $x_i \to y_i$ is used to replace the leftmost occurrence of $x_i$ by $y_i$, therefore transforming $z$ into $z'$. The process continues until a special rule $x\to \cdot$ is encountered, indicating that the algorithm terminates and returns the transformed value of $z$, or the process is blocked. Appendix~\ref{app:markov} provides examples of Markov algorithms.

Markov algorithms can be shown to be Turing-complete: any finite computation can be implemented by Markov algorithms, which therefore constitute a complete model of computation. Any language model that can be trained to implement rewrite rules can serve as a universal computation tool. In this paper, we consider two tasks:
\begin{itemize}[noitemsep,nolistsep]
    \item learn to apply the rule on a sequence where it is applicable
    \item learn to apply the rule if it is applicable or return the input sequence 
\end{itemize}

The first task is the basic rewrite operation. The second allows the algorithm to move to the next rule if the current one does not apply.

\begin{figure*}
     \centering
     \begin{subfigure}[h]{.38\textwidth}
         \centering
         \includegraphics[width=\columnwidth]{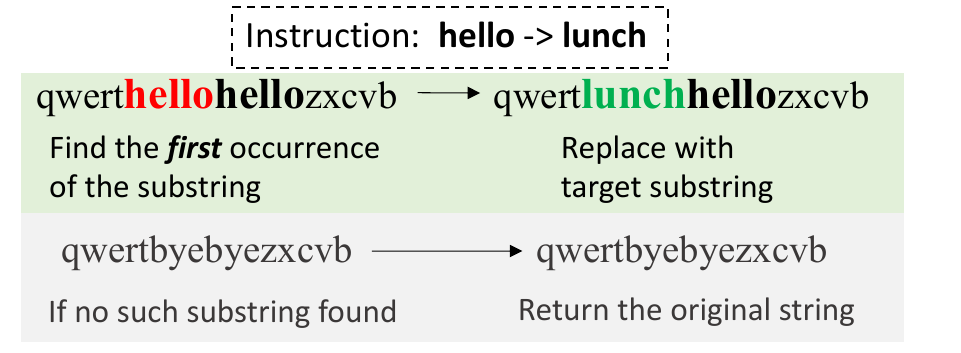}
         \caption{Our string-rewriting task set-up.}
         \label{fig:markov}
     \end{subfigure}
     \centering
     \begin{subfigure}[h]{.6\textwidth}
         \centering
         \includegraphics[width=\columnwidth]{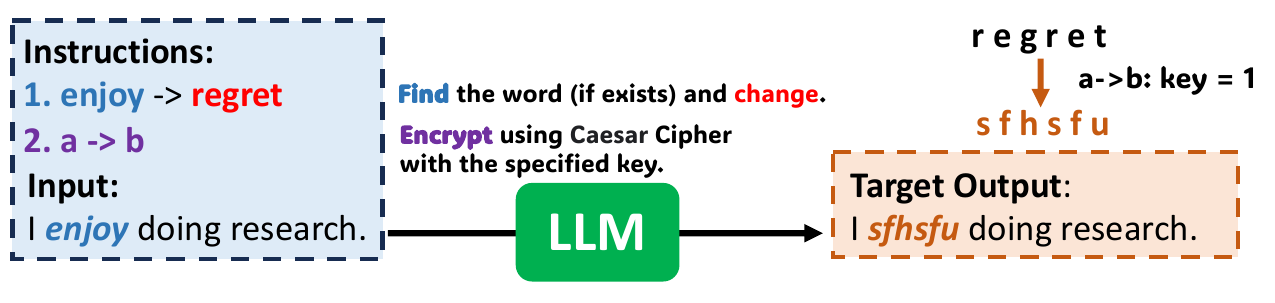}
         \caption{Encrypted Re-writing. Similar to the re-writing experiment, but the model needs to infer the rule of encryption based on the rule of shifting specified in the instructions and compute the encrypted word.}
         \label{fig:realworld}
     \end{subfigure}
     \caption{Illustration of our symbolic tasks in this paper.}
\end{figure*}



All the strings considered in our experiments are sequences of the lowercase Latin letters $a\dots z$. Model inputs are triplets of strings, $(x,y,z)$, representing the rule $x \to y$ and the input sequence $z$. Model outputs are obtained by replacing the leftmost instance of $x$ by $y$, in the sequence $z$. If $x$ does not appear in $z$, the model output is $z$ (input sentence is copied, Figure~\ref{fig:markov}).




\section{Experiment Results}

\subsection{Instruction Diversity Drives Generalization}
\begin{figure*}
     \centering
     \begin{subfigure}[t]{.44\textwidth}
         \centering
         \includegraphics[width=\columnwidth]{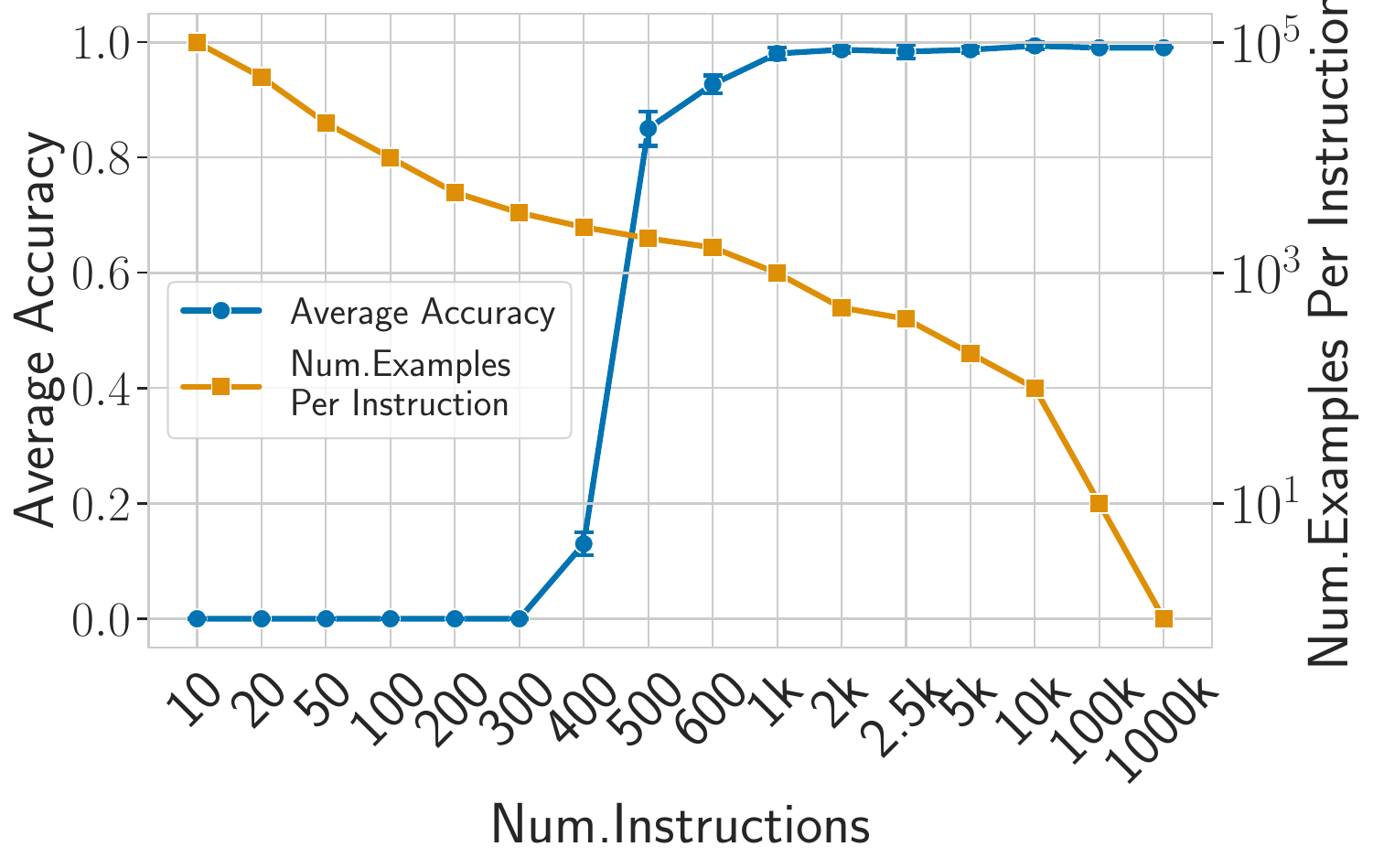}
         \caption{Re-writing accuracy against the number of instructions with a fixed-size training set.}
         \label{fig:one_occurence}
     \end{subfigure}
     \centering
     \begin{subfigure}[t]{.42\textwidth}
         \centering
         \includegraphics[width=\columnwidth]{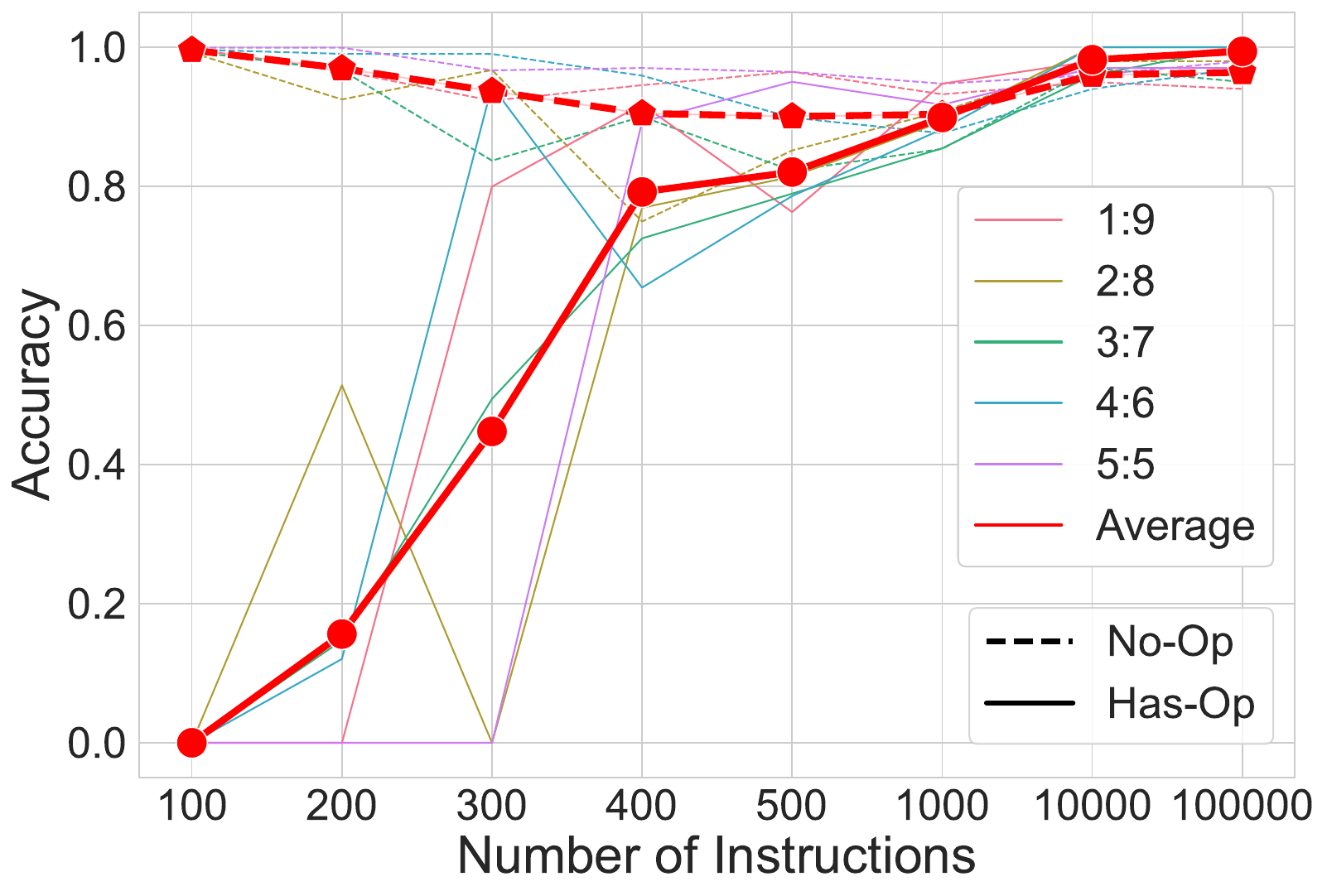}
         \vfill
         \caption{Rewriting with no-op situation included. }
         
         \label{fig:zero_one}
     \end{subfigure}
     \caption{Generalization versus number of instructions during training. }
\end{figure*}

In the first set of experiments, we train GPT-2 models with 6 layers, 4 heads, and hidden dimension 256 (see Appendix~\ref{app:input_semantics} for more information on our experimental setting) on a generated sample of $S\times I$ input sequences, corresponding to $I$ different replacement rules (instructions) applied to $S$ different sequences. The trained model is then tested on a dataset of $10^5$ examples with unseen instructions. Figure~\ref{fig:one_occurence} presents the generalization accuracy for models trained on $10^6$ examples as a function of the number of different instructions in the training set. We note that models trained on less than $300$ instructions never generalize, even when the model only has a few rules to learn and is provided with a very large number of examples per rule. On the other hand, models trained on $1,000$ instructions or more always generalize, even when the number of instructions becomes very large, and each rule is only featured in a handful of examples. A very sharp phase transition happens around $400$ instructions. 

We conclude that the number of instructions in the training set ($I$) is the key factor that allows the model to generalize to different instructions and variation in input semantics (see Appendix~\ref{app:input_semantics}) instead of $S$.


\subsection{Searching and replacing, the roles of No-Ops}
\label{sec:no_op}
So far, the sub-string to be replaced was always featured in the input sequence. We now consider the more general case where some rules do not apply. In such cases, which we call ``no-ops'', the model returns its original input. The model must now learn a two-step task: check whether a rule is a no-ops, then apply it. This is closer to a real-world setting, where replacements often do not apply. It also corresponds to the ``skip inapplicable rule and try the next one'' case in Markov algorithms. We evaluate model generalization to unseen instructions as a function of the number of instructions in the training set and the frequency of no-ops, which we vary between $50\%$ and $10\%$ (Figure~\ref{fig:zero_one}). The size of the training and test sets are the same as in previous experiments.

On average, the addition of no-ops has little impact on model performance for ``has-ops'' cases (where replacement happens): generalization only happens in models trained on more than $400$ different instructions (vs $500$ previously). No-ops cases, on the other hand, are always correctly predicted, and models trained on a few different instructions default to always predicting no-ops. The frequency of no-ops in the training sample has a marginal impact on performance: our general conclusions remain the same, but the number of instructions needed for generalization is slightly lower when there are fewer no-ops.


 
\begin{figure*}
     \centering
     \begin{subfigure}[h]{.5\textwidth}
         \centering
         \includegraphics[width=.8\columnwidth]{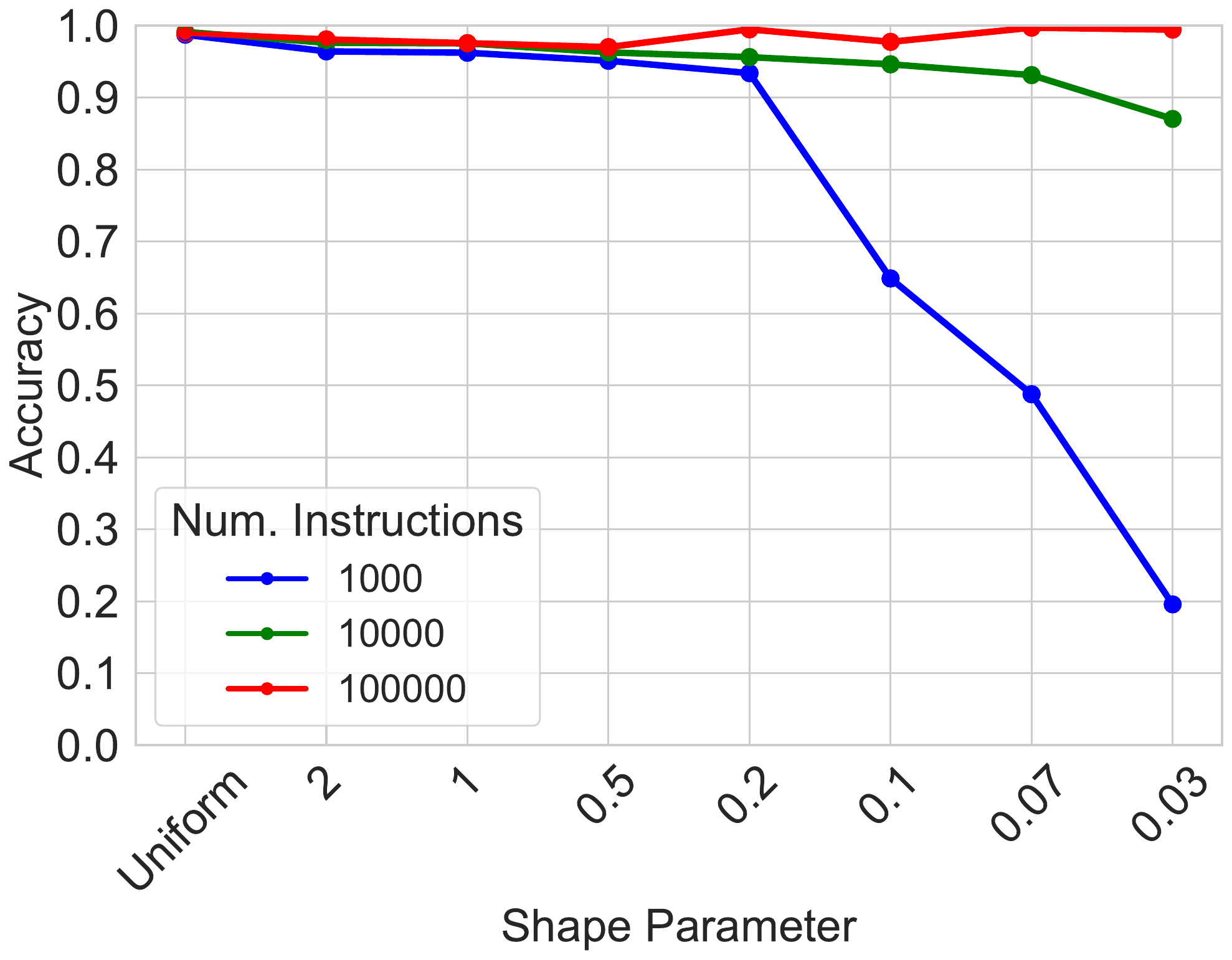}
         \caption{Effect of long-tail task distributions on model's generalization ability.}
         \label{fig:long_tail_distribution}
     \end{subfigure}
     \centering
     \begin{subfigure}[h]{.42\textwidth}
     \centering
         \includegraphics[width=.9\columnwidth]{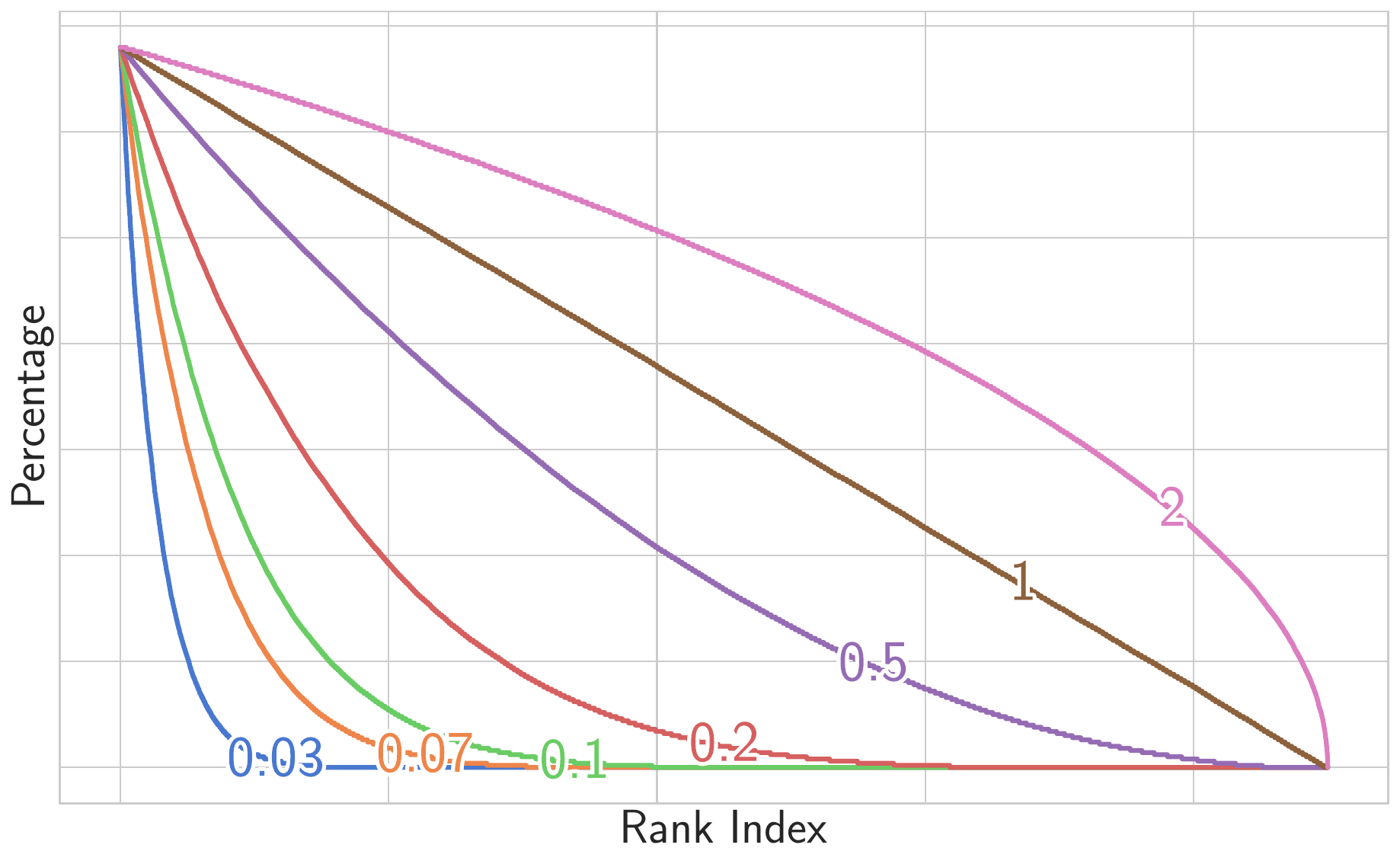}
         \caption{The sorted percentage of each instruction following power-law distribution with different shape parameters. The y-axis is the percentage of the rules in the training mixture. The x-axis is the ranked index (by proportion of examples) of instructions.}
         \label{fig:distribution}
     \end{subfigure}
     \caption{Generalization versus number of instructions during training. }
\end{figure*}

We use power-law distribution in the distribution experiment in Section ~\ref{sec:rule_distribution}. Figure~\ref{fig:distribution} is a visualization of how the percentages of rule decay. 


\subsection{Diversity of Instruction Semantics}

Previous experiments demonstrate the merit of training models on a large set of instructions. We now investigate the impact of semantic diversity in the instruction set. To this effect, we train models on a large but restricted set of rules by constraining the sub-strings to replace or to be replaced and test them on less constrained instructions. We experiment with three sets of constrained substrings:
\begin{itemize}[noitemsep,nolistsep]
\item characters repeated $k$ times: $aaabbbccc$ for $k=3$
\item patterns repeated $k$ times: $abcabc$ for $k=2$
\item mirroring patterns repeated $k$ times: $abccbaabc$ for $k=3$.
\end{itemize}

\begin{figure}
     \centering
         \includegraphics[width=\columnwidth]{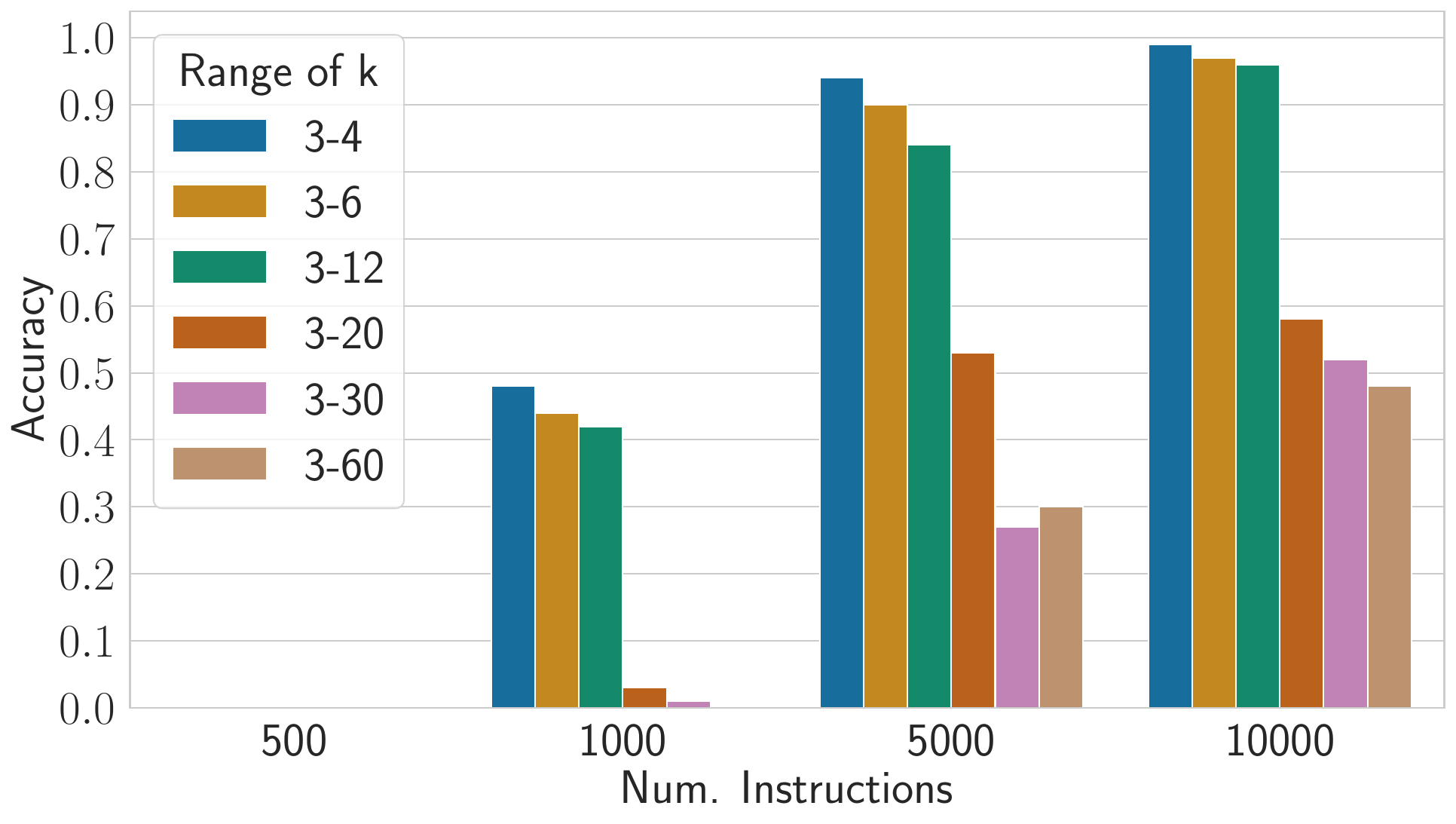}
         \caption{Model's performance when trained on the three classes of restricted semantics. Models trained on 500 or less instructions never generalize.}
         \label{fig:semantics}
\end{figure}
In all three settings, large values of $k$ correspond to more constrained instructions. We train models on instructions with large $k$ and test them on low $k$. We observe that models trained on one set of constrained (repeated, periodic, or mirror) do not generalize to lower $k$, and mixing repeated and periodic instructions does not help. Although the model can generalize to unseen tasks in-domain (same $k$), the accuracy is always zero on lower $k$ (or unrestricted). 

The situation changes when the three training sets are mixed together (Figure ~\ref{fig:semantics}). Models trained on large $k$ (for all three constraints) do generalize to small $k$ (and other unconstrained instructions). As before, a larger number of instructions in the training set improves model generalization. Also, the more constrained the training instructions are (i.e. larger values of $k$), the harder generalization becomes.

Overall, we notice that just as training on a large set of instructions is key to achieving generalization to unseen instruction, training on several semantically constrained sets of rules allows for generalization to semantically unconstrained rules.

\subsection{Impact of instruction Distributions}
\label{sec:rule_distribution}
In all previous experiments, training examples were evenly distributed between instructions. This situation can be achieved in controlled experiments, but it is unlikely to happen in real-world scenarios. We now investigate the impact of unbalanced instruction training sets. To this effect, we fix the size of the training set and the number of different instructions constant but distribute the number of examples for each instruction according to a power law and experiment with different shape parameters, ranging from uniform laws to situations where a few training instructions are very common, but most are extremely rare.




Figure~\ref{fig:long_tail_distribution} presents model generalization as a function of the variance of the distribution of examples for training sets with 1,000, 10,000, and 100,000 instructions. For models trained on $1000$ instructions (the minimal diversity level to guarantee generalization, according to our previous experiments), performance drops steeply once the example distribution becomes too uneven. Models trained on larger instruction sets, on the other hand, suffer little penalty. This confirms our earlier observations about the phase transition happening around $500$ examples. A long-tailed example distribution amounts to a reduction in the number of training instructions. It impacts models trained on limited sets of instructions -- close to the cut-off limit of $500$.

\subsection{Generalization Across Input Semantics}
\label{app:input_semantics}
To assess the impact of instruction diversity on a model's ability to generalize across a broader semantic input space, our research explored how well the model could identify and manipulate specific sub-string occurrences within sequences previously during its training phase. We specifically trained the model on datasets characterized by limited occurrence frequencies and then evaluated its performance across a spectrum ranging from 1 to 20 occurrences.

The findings, as shown in Table~\ref{tab:input_occurrence}, reveal a marked decline in performance when the model is confronted with variable occurrence counts compared with situations with a consistent single occurrence. This variability tends to confound the model, impeding its ability to recognize the input strings accurately and, consequently, to apply the instructions as intended.

Crucially, however, enriching the training set with a wider array of occurrence frequencies demonstrably enhances the model's performance. It becomes evident that incorporating examples containing occurrence numbers from each sub-interval within the overall range significantly boosts the model's operational effectiveness.

The key takeaway from our study is the undeniable importance of instruction diversity. A training regimen infused with various instructions substantially improves the model's capability to generalize to unseen occurrences. This enhanced generalization ability underscores the critical role that diverse instruction plays in making the model generalize to unseen semantic input spaces more effectively.

\begin{table}[]
\begin{tabular}{@{}l|llll@{}}
\toprule
Train Occ. \textbf{\textbackslash} Num. Inst & 2000 & 1000 & 500  & 200  \\ \midrule
1                                                & 0.71 & 0.41 & 0.00 & 0.00 \\
10                                               & 0.88 & 0.71 & 0.00 & 0.00 \\
15                                               & 0.84 & 0.59 & 0.00 & 0.00 \\
20                                               & 0.73 & 0.28 & 0.07 & 0.00 \\
1,5,10,15,20                                     & 0.94 & 0.94 & 0.62 & 0.00 \\ \bottomrule
\end{tabular}
\caption{Results on the generalizability across input semantics. Train Occ. Denotes the occurrences of the substring in the training dataset, Num. Inst denotes the number of diverse instructions.}
\label{tab:input_occurrence}
\end{table}
\section{Experiments with Pre-Trained Models}
All models so far were trained from scratch. In this section, we show how our experiments can be extended to pre-trained models. LLM already learns basic string replacement during pre-training, so we need a more difficult task.  
We introduce an encrypted-rewriting task that requires multi-hop reasoning (Figure~\ref{fig:realworld}). The task involves a sentence and an encrypted re-writing instruction. The model first replaces the specified word with another word, then encrypts that word using Caesar cipher with a key specified in the instruction.

\begin{figure}
     \centering
         \includegraphics[width=\columnwidth]{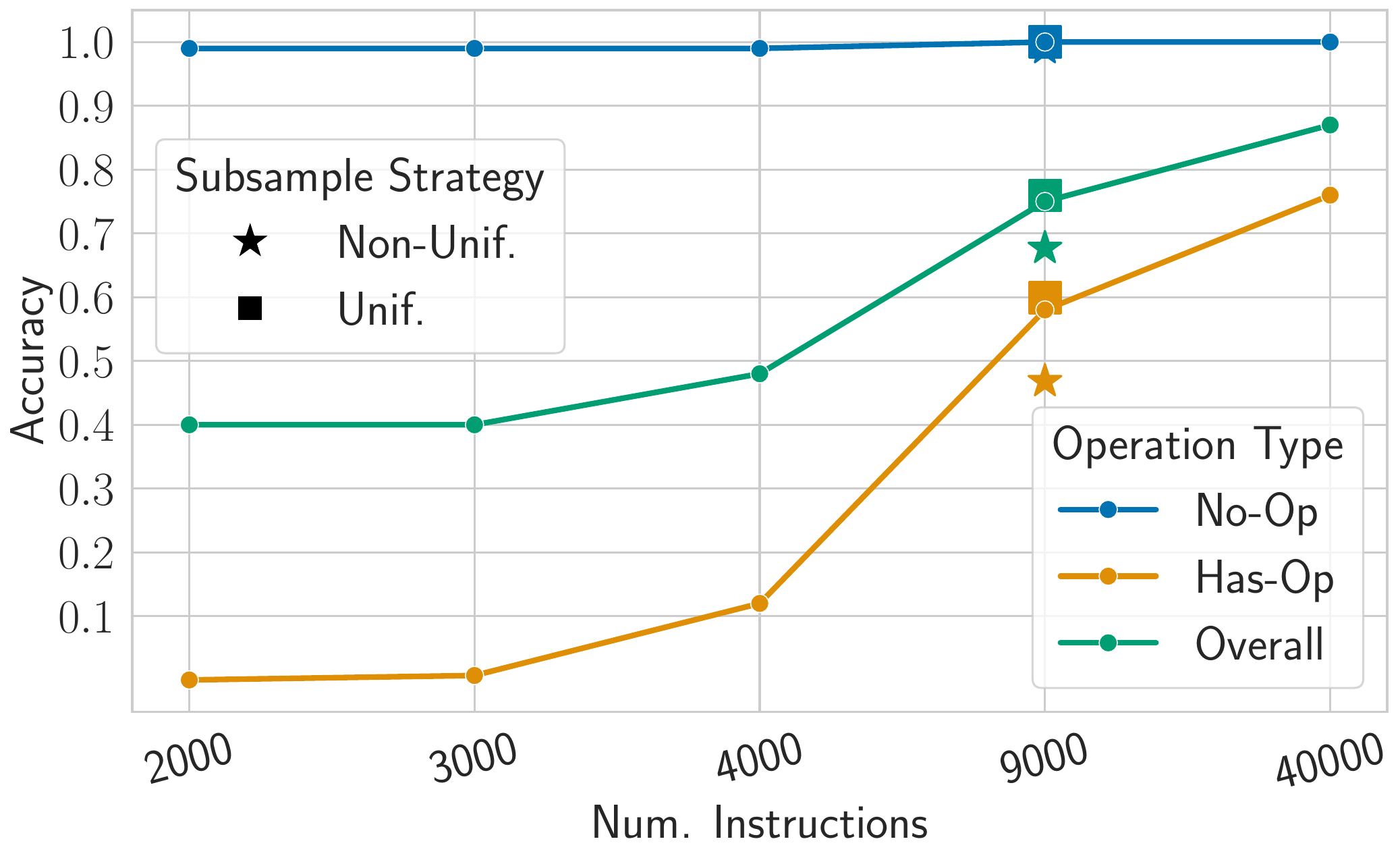}
         \caption{Performance of Llama-2 model on the encrypted-rewriting task. We also conducted uniform / non-uniform sub-samplings to half the total sample size at 9000 instructions. Uniform sub-sampling does not harm performance whereas non-uniform subsampling impacts generalization.}
         \label{fig:realworld_perf}
\end{figure}

We keep two disjoint dictionaries for train and test sets and prompt GPT-3.5-turbo to generate sentences containing words from the dictionary. If the word is in the generated sentence, we randomly sample a replacement and encrypt it with a random key. In no-ops cases, the input should be returned. We generate training sets of $40,000$ sequences and test them on sets of $5,000$ instances - each generated using a distinct word in the test dictionary and again a randomly chosen key. Both sets contain $40\%$ no-ops cases.

We fine-tuned the pre-trained language model (Llama2-7b)~\cite{touvron2023llama} with LoRA~\cite{hu2021lora} with rank 512 and $\alpha$ of 1024 till convergence.  Consistent with our earlier observations, the diversity of instructions benefits the model's generalization. With a smaller number of instructions, the pre-trained LLM also only solves no-op cases but cannot correctly perform the ``replace-then-encrypt'' operation (Figure~\ref{fig:realworld_perf}).

\section{Conclusion}
Through our symbolic experiments, we have shown that language models only generalize to instructions unseen during training when trained on a large and diverse set of instructions. For a fixed data budget, instruction diversity outweighs better illustration (i.e. more examples) for each instruction. These observations apply not only to the number of different instructions in the dataset but also to their semantic diversity. The negative effect of an unbalanced distribution of examples can be counteracted by a larger number of instructions in the training set. We also empirically demonstrated that these results indeed apply to the fine-tuning of pre-trained language models.


\section*{Limitations}
\textbf{Lack of real-world instruction datasets.} To gain full control over different factors and ablate the effect of each, we adopted a synthetic setup instead of experimenting with real-world instruction-following datasets. The abstraction might ignore certain attributes like knowledge during pre-training, language, etc. That being said, these factors are not the main focus of this work. 

\textbf{Lack of theoretical justifications.} The results are empirical. Our future work shall seek theoretical justifications for the conclusions we had. 
\section*{Ethics Statement}

Instruction tuning may have ethical consequences. If the fine-tuning data is not controlled or has low quality, biases can be introduced as the model is fine-tuned. We believe our results suggest a possible mitigation to such biases. By increasing the diversity of instructions in fine-tuning sets, models learn to generalize to unseen distributions and may therefore be less susceptible to biases in their training data. Additional research and extensive experiments would be needed to turn this into a practical technique.

\bibliography{anthology,main}
\bibliographystyle{acl_natbib}
\newpage
\appendix


\section{Complement on Markov algorithms}
\label{app:markov}

Markov algorithms~\cite{markov54} are ordered sets of rewrite rules, operating on sequences of symbols in a fixed alphabet $\mathcal U$. A sequence $S$ is processed by applying the first rewrite applicable to $S$, at the leftmost position if several exist: i.e. the rewrite rule $ss \to tr$ transforms the sequence $S=mississipi$ into $S'=mitrissipi$. The algorithm is then applied to $S'$, and the process is repeated until either no rules apply, and the algorithm is said to be \emph{blocked}, or a special rule, called a \emph{stop rule} is invoked, and the algorithm terminates and returns the final rewritten sequence.

Specifically, the algorithm uses and alphabet $\mathcal A$, which includes the alphabet $\mathcal U$ used buy the sequences to be processed (henceforth, small case latin letters), a set of additional symbols (henceforth, the small case greek letters $\{\alpha, \beta \dots \}$, and a special symbol $\cdot$ indicating a stop rule.

For instance, we could define the following algorithm, with $\mathcal U=\{a,b\}$, and $\mathcal A=\{a,b,\alpha,\beta,\cdot\}$, and the rules

\begin{eqnarray} 
\alpha x &\to& x \alpha \beta x \\
\beta xy &\to& y \beta x \\
\alpha \beta x &\to& x \alpha \\
\alpha &\to& \cdot \\
 &\to& \alpha
\end{eqnarray}
where $x$ and $y$ stand for any letter $a$ or $b$. This will transform any sequence of $a$ and $b$ into a concatenation of the sequence and its reverse. Applied on $abb$, the algorithm will perform the following rewrites: 

\begin{align*} 
abb &\to \alpha abb &&(\text{by }5)\\
\alpha abb  &\to a \alpha \beta abb &&(\text{by }1)\\ 
a \alpha \beta abb &\to a \alpha b \beta ab &&(\text{by }2)\\
a \alpha b \beta ab &\to a b \alpha \beta b \beta ab &&(\text{by }1)\\
 a b \alpha b \beta b \beta ab &\to  a b \alpha \beta bb \beta a &&(\text{by }2)\\
 a b \alpha \beta bb \beta a&\to a b \alpha b \beta b \beta a &&(\text{by }2)\\
a b \alpha b \beta b \beta a &\to abb \alpha \beta b \beta b \beta a &&(\text{by }1)\\ 
abb \alpha \beta b \beta b \beta a &\to abb b \alpha \beta b \beta a &&(\text{by }3)\\
abb b \alpha \beta b \beta a &\to abbbb \alpha \beta a &&(\text{by }3)\\ 
abbbb \alpha \beta a &\to abbbba \alpha &&(\text{by }3)\\
abbbba \alpha &\to abbbba &&(\text{by }4)
\end{align*}

Since rule $4$ is a stop rule, the algorithm terminates and returns $abbbba$.

Judicious introduction of additional (greek) letters allows one to compose Markov algorithms, effectively writing complex programs. Any effective process (i.e. finite computation) can be represented as a Markov algorithm (this is Markov's thesis).

\section{Experimental set-up}
\label{task}
\subsection{Model and Training}
In rewrite experiments, we train GPT-2 models~\cite{Radford2019gpt2}, a decoder-only transformer-based architecture, with $6$ layers, $256$ dimensions and $4$ attention heads from scratch, on a generated instruction-tuning dataset using standard supervised fine-tuning approach.  We use the AdamW optimizer, a learning rate of $10^{-3}$, and linear scheduling. All models are trained for 50 epochs. For the encrypted-rewriting task, we LoRA fine-tuned Llama-2 models with a learning rate of 1e-4, batch size 64, 8-bit quantization. The model takes about 2000 steps to converge. We used greedy decoding for all experiments.

\subsection{Data Generation}

Except for the diversity of semantics experiment, the results we reported in the main paper are obtained from an input length of 50 and a pattern length of 20.
To validate the generality of our findings, we conducted experiments on various input sizes \{50, 100, 200\} and, correspondingly, pattern lengths \{20,40,50\}.

In the diversity of semantics experiment, we used an input length of 500 and a pattern length of 60. We strictly restricted the sub-strings to look for and to replace them with both to be unseen during testing.

\end{document}